\documentclass[conference]{IEEEtran}
\IEEEoverridecommandlockouts
\def\BibTeX{{\rm B\kern-.05em{\sc i\kern-.025em b}\kern-.08em
 T\kern-.1667em\lower.7ex\hbox{E}\kern-.125emX}}

\usepackage[margin=10pt,font=small,labelfont=bf,justification=justified]{caption}
\usepackage[legalpaper, margin=1in]{geometry}
\usepackage[dvipsnames]{xcolor}

\usepackage{subcaption,url}
\usepackage[noadjust]{cite}
\usepackage{mathtools, slashbox,lipsum,dsfont,yhmath,accents,footmisc,comment,shadow}
\usepackage{amsmath,amssymb,mathrsfs,amsfonts,amsthm,soul}
\usepackage{enumerate}
\usepackage[english]{babel}
\usepackage{epsfig,amsfonts,latexsym,amssymb,amsmath,wrapfig}
\usepackage{tikz}
\usepackage{algorithm}
\usepackage{algpseudocode}
\usepackage{algorithm}
\usepackage{footnote}
\usepackage{tablefootnote}
\makesavenoteenv{tabular}

\newtheorem{remark}{Remark}

\newtheorem{assum}{Assumption}

\DeclareMathOperator*{\argmax}{\arg\!\max}

\newcommand{\blue}[1]{\textcolor{blue}{#1}}

\newcommand{\yellow}[1]{\textcolor{yellow}{#1}}
\newcommand{\red}[1]{\textcolor{red}{#1}}

\newcommand{\ubar}[1]{\underaccent{\bar}{#1}}

\newcounter{l1}
\newcounter{l2}
\newcounter{l3}
\newcommand{\bdotlist}{\begin{list}{$\bullet$}{}}
\newcommand{\bboxlist}{\begin{list}{$\Box$}{}}
\newcommand{\bbboxlist}{\begin{list}{\raisebox{.005in}{{\tiny $\blacksquare$ \ \ }}}{}}
\newcommand{\bdashlist}{\begin{list}{$-$}{} }
\newcommand{\blist}{\begin{list}{}{} }
\newcommand{\barablist}{\begin{list}{\arabic{l1}}{\usecounter{l1}}}
\newcommand{\balphlist}{\begin{list}{(\alph{l2})}{\usecounter{l2}}}
\newcommand{\bAlphlist}{\begin{list}{\Alph{l2}.}{\usecounter{l2}}}
\newcommand{\bdiamlist}{\begin{list}{$\diamond$}{}}
\newcommand{\bromalist}{\begin{list}{(\roman{l3})}{\usecounter{l3}}}

\begin{document}

\title{A Generative Machine Learning Approach to Policy Optimization in Pursuit-Evasion Games}

\author{\IEEEauthorblockN{1\textsuperscript{st} Shiva Navabi}
\IEEEauthorblockA{\textit{Electrical and Computer Engineering Department} \\
\textit{University of Southern California}\\
Los Angeles, USA \\
navabiso@usc.edu}
\and
\IEEEauthorblockN{2\textsuperscript{nd} Osonde A. Osoba}
\IEEEauthorblockA{\textit{RAND Corporation} \\
Santa Monica, USA \\
oosoba@prgs.edu}
}

\maketitle

\begin{abstract}
We consider a pursuit-evasion game \cite{isaacs1999differential} played between two agents, `Blue' (the pursuer) and `Red' (the evader), over $T$ time steps. 
Red aims to attack Blue's territory. 
Blue's objective is to intercept Red by time $T$ and thereby limit the success of Red's attack.
Blue must plan its pursuit trajectory by choosing parameters that determine its course of movement (speed and angle in our setup) such that it intercepts Red by time $T$.
We show that Blue's path-planning problem in pursuing Red, can be posed as a sequential decision making problem under uncertainty. 
Blue's unawareness of Red's action policy renders the analytic dynamic programming approach intractable for finding the optimal action policy for Blue.
In this work, we are interested in exploring data-driven approaches to the policy optimization problem that Blue faces.
We apply generative machine learning (ML) approaches to learn optimal action policies for Blue. 
This highlights the ability of generative ML model to learn the relevant implicit representations for the dynamics of simulated pursuit-evasion games.
We demonstrate the effectiveness of our modeling approach via extensive statistical assessments.
This work can be viewed as a preliminary step towards further adoption of generative modeling approaches for addressing policy optimization problems that arise in the context of multi-agent learning and planning \cite{albrecht2018autonomous}. 

\end{abstract}

\section{Introduction}
We consider the problem of optimizing an agent's action policy for achieving a desired objective in a multi-stage game against an opponent. 
We frame the problem as a pursuit-evasion game \cite{isaacs1999differential} where an agent (Blue) chases its enemy (Red) with the objective of catching Red before it reaches a critical region in Blue's territory. 
Such scenarios can arise in practice in aircraft pursuit interactions. 
Blue needs to judiciously decide on its motion-associated parameters (such as speed, orientation, \& acceleration) in order to catch the enemy in a timely fashion.
The strategic nature of the Red agent imposes a further level of complexity. 
The Blue agent needs to dynamically monitor Red's course of movement and adjust its path accordingly so as to stand a reasonable chance of neutralizing Red's invasion. 
Blue's decision problem also needs to account for inherent physical constraints (like limited fuel availability) when planning its path.

\subsection{Framing the Decision Problem}\label{subsec:math-set}
Let $\mathcal{A}$ denote the set of admissible actions (i.e., the admissible motion-associated parameter values) available to Blue. 
From a game-theoretic standpoint, Blue's decision problem can be formulated as finding the \textit{best response} against Red in each round of the $n$-stage game:
\begin{align}
a^*_i := \argmax\limits_{a \in \mathcal{A}} \; Q_i(a,h_i) \; , \; i = 1, \ldots, n \; , \label{eq:BlueProb}
\end{align}
where $a^*_i$ represents the optimal action for Blue in the $i$th stage of the game. 
$a^*_i$ (in (\ref{eq:BlueProb})) is the maximizer of the value function $Q_i(\cdot)$ which depends on Blue's action $a$ as well as Blue's information in the $i$th stage denoted as $h_i$. 
$h_i$ includes Blue's observations of Red's chosen actions in the unfolded stages of the game. 
In a pursuit-evasion game setup $Q_i(a,h_i)$ may indicate the effectiveness of Blue's adopted movement trajectory (encoded in $a$) in thwarting Red's invasion in a timely fashion. 
For instance, $Q_i(a,h_i)$ could be a measure of the ultimate spatial proximity between Blue and Red in the terminal stage of the pursuit-evasion game, when given the information $h_i$ Blue chooses action $a$ in stage $i$. 

We are interested in developing a computational model for the optimal action policy for Blue's decision problem in each stage of the game. 
Let $\pi^*_i(\cdot)$ denote the optimal \textit{stochastic} action policy for Blue's decision problem in the $i$th stage of the game. 
Given $h_i$, $\pi^*_i(h_i)$ outputs the optimal action  for Blue in response to Red's actions as observed by Blue (included in $h_i$). 
In this work, we consider addressing a \textit{probabilistic relaxation} of Blue's policy optimization problem. 
More precisely, given some $h_i$ in the $i$th stage of the game we seek to find the stochastic action policy $\pi^*_i(\cdot)$ such that for any small $\epsilon > 0$
\begin{align}\label{eq:probRelax}
\mathbb{P}\Big( | Q_i(\pi^*_i(h_i) , h_i) - Q_i(a^*_i, h_i)| > \epsilon \Big) \approx 0. 
\end{align} 
That is, we aim to find the stochastic action policy $\pi^*_i(\cdot)$ such that given any $h_i$, it would optimize the value function $Q_i(\cdot)$ \textit{with high probability}. 
Focusing on the probabilistic relaxation in (\ref{eq:probRelax}) expands the set of applicable models that can suitably represent Blue's action policy. 
In particular, the generative modeling frameworks such as Generative Adversarial Networks (GANs) \cite{goodfellow2014generative} and Variational Autoencoders (VAEs) \cite{kingma2013auto, rezende2014stochastic} developed under the generative machine learning paradigm \cite{bengio2009learning} seem to be a great fit for this purpose. 
Given that generative models can provide implicit representations of the probability distributions of interest, we can use them to represent Blue's (stochastic) action policy. 
The trained generative model can be efficiently queried for samples from the target stochastic policy (i.e., $\pi^*_i(\cdot)$ in our setup) in response to a Red challenge.
The policy responses can then be filtered or ranked to approach optimality. \\
We can also rely on Monte Carlo theories and methods~\cite{kahn1954applications, mohamed2019monte} to guarantee useful approximations of the quality of the policy responses.
In principle, this approach enables us to find the best response in any instance of Blue's decision problem in (\ref{eq:BlueProb}) \textit{with high probability} and rule out sub-optimal or even unfavorable actions that may be recommended by the generative model with non-zero probability. 


The work reported in this paper shows how we construct and optimize GAN-based decision architectures and deploy them in a sequential fashion to address Blue's decision problem in the multi-stage game against an adversarial agent (Red). 
We provide results from extensive simulation experiments based on a synthetic two-stage pursuit-evasion game setup to demonstrate effectiveness of our proposed decision model. 
Our work can be viewed as a preliminary step towards further applications of deep generative modeling approaches \cite{kingma2014semi} in simultaneously addressing the opponent modeling \cite{carmel1995opponent} and policy optimization \cite{sutton2018reinforcement} problems that are long-standing challenges in multi-agent systems. 
Indeed extensive research is required to address more complicated aspects of the multi-agent decision problems including \textit{non-stationarity} of the agents \cite{hernandez2017survey}. 

\begin{figure*}[!ht]
	\centering
	\includegraphics[width=2\columnwidth, keepaspectratio]{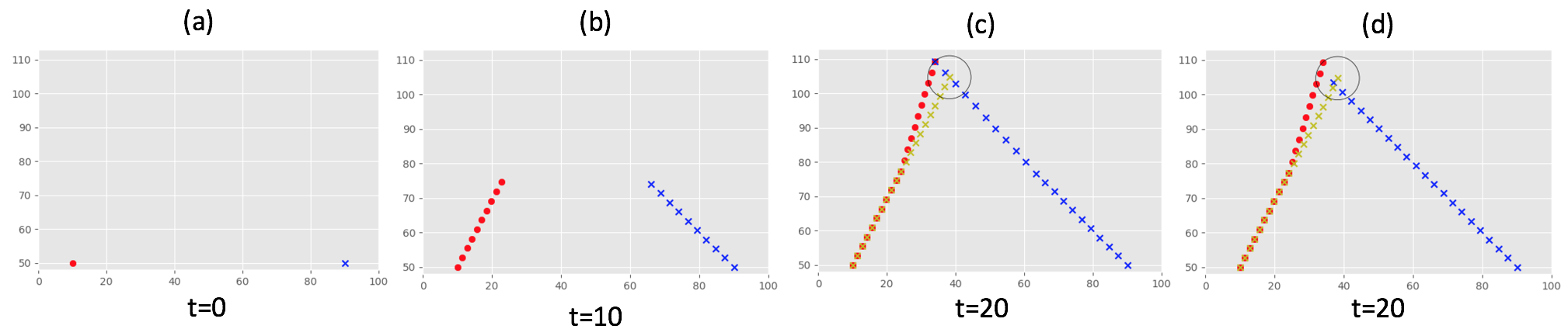}
	\caption{An instance of the two-stage pursuit-evasion game displayed in three snapshots at times $t=0,10,20$. (a) just depicts the initial locations of Red and Blue at time $t=0$: Red starts its path from $(X^R_0,Y^R_0) = (10,50)$ (marked by \red{$\bullet$}) while Blue starts its path from $(X^B_0,Y^B_0) = (90,50)$ (marked by \blue{$\pmb{\times}$}). (b) depicts Red's and Blue's paths traveled during the first stage of the game, i.e., from time $t=0$ up until $t=11$. Each \red{$\bullet$} and \blue{$\pmb{\times}$} marks Red's and Blue's locations, respectively, at the end of the corresponding time step on their associated paths. Red initiates the game at time $t=0$ by moving from its initial location towards some target on Blue's territory (see (b)). Blue starts chasing Red from its initial location $(X^B_0,Y^B_0)$ at time $t=0$. (c) shows the full trajectories of Red's and Blue's movements as unfolded at termination of the two-stage game, i.e., time $t=20$. The game instance illustrated in (a)-(c) represents a \textit{successful} chase maneuver for Blue, i.e., one where Blue's path coincides with Red's path on Red's safety circle delineated in the (c), exactly at time $t=20$, without violating the speed upper bound $\overline{V}$, i.e., Blue's speed parameters satisfy $V^B_1 + V^B_2 \le \overline{V}$. The yellow crosses \yellow{$\pmb{\times}$} in (c) mark Red's path as \textit{initially projected} by Blue, had it not changed its course of movement in the second half of the game. As can be seen in (c), at time $t=11$ Red deviates from the yellow line due to switching to a new path governed by a new pair of speed and angle parameters decided based on Red's observation of Blue's location at that time. (d) depicts the pursuit-evasion game with the exact same Red's path in stage 1 as in (b), in which Blue's entire path is planned based on the parameters $(V,\Theta)$ that it decides after observing $(V^R_1,\Theta^R_1)$ at time $t=0$. The initial locations of Red and Blue at time $t=0$ are the same as in (a). While Blue's trajectory emerges as a single line, Red still responds to its observation of Blue's location at time $t=11$ and updates its path accordingly. As can be seen in (d), at time $t=11$ Red deviates from the yellow line toward its updated destination on its safety circle. Remaining on its initially decided path, Blue therefore ends up near the point it initially had projected Red to get to by $t=20$, i.e., the end-point of the yellow line. }
	\label{fig:trajs3step}
\end{figure*}

\section{Problem Setup}\label{sec:setup}
\textit{Notation:} $x_{1:n}$ is the short-hand for the vector $(x_1, \ldots, x_n)$. $\Delta(\mathcal{A})$ denotes the space of all probability distributions with the support equal to the set $\mathcal{A}$. $\mathbb{P}$ denotes the probability measure. Random variables are denoted by upper case letters (e.g., $V, R, B, X, Y, \Theta$), their realizations by the corresponding lower case letters (e.g., $v, r, b, x, y, \theta$). $\mathcal{N}(\mu,\sigma)$ denotes the Gaussian probability distribution with mean value $\mu$ and standard deviation $\sigma$.

Consider a pursuit-evasion game \cite{isaacs1999differential} between two players: `Blue' (the pursuer) and `Red' (the evader). Blue attempts to protect its territory against Red's attack. The game is played over $T$ time steps in two equally long consecutive stages. At time $t=0$, Red initiates the game by moving towards a target on Blue's territory from its initial location $(X^R_0,Y^R_0)$ on the 2D plane. Red's movement within the first $\frac{T}{2}$ time steps (stage 1) is modeled as a monotonous motion on the 2D plane, governed by the initial speed $V^R_1$ and initial angle $\Theta^R_1$ parameters that are decided by Red at time $t=0$. Right after Red starts moving, Blue observes the pair of parameters $(V^R_1,\Theta^R_1)$ and based on that, starts a chase towards the point it \textit{projects} Red to end up after moving for $T$ time steps in accordance with the speed and angle pair $(V^R_1,\Theta^R_1)$. 
Let $(\tilde{X}^R_{T}, \tilde{Y}^R_{T})$ denote Red's destination as \textit{initially} projected by Blue based on the observed parameters $(V^R_1,\Theta^R_1)$. 

Blue starts  chasing Red from its fixed initial location $(X^B_0,Y^B_0)$. Blue's chase is modeled as a monotonous motion on the 2D plane, governed by the initial speed $V^B_1$ and the initial angle $\Theta_1^B$ that are decided by Blue such that it can catch Red at time $t=T$ at the projected destination $(\tilde{X}^R_{T}, \tilde{Y}^R_{T})$. 

At the end of time $t=\frac{T}{2}$, Red observes Blue's current location $(X_{\frac{T}{2}}^B,Y_{\frac{T}{2}}^B)$ and adjusts its path so as to reach a new destination that meets the following two criteria:
\begin{itemize}
\item Red ends up on its \textit{safety zone} which is delineated by the perimeter of a circle centered at $(\tilde{X}^R_{T}, \tilde{Y}^R_{T})$ with radius equal to $\delta\%$ of the distance between $(X^R_0,Y^R_0)$ and $(\tilde{X}^R_{T}, \tilde{Y}^R_{T})$.
\item Red reaches the farthest from Blue's location at time $t=\frac{T}{2}$, i.e., $(X_{\frac{T}{2}}^B,Y_{\frac{T}{2}}^B)$, while remaining on its safety circle. 
\end{itemize}
Red's movement within times $t = \frac{T}{2}+1, \ldots, T$ is governed by the new pair of parameters $(V^R_2,\Theta^R_2)$ that are chosen such that the above two criteria are satisfied. 

At time $t=\frac{T}{2}$ Blue notices the alteration in Red's movement trajectory through observing the new pair $(V^R_2,\Theta^R_2)$. Blue then updates its prediction about Red's destination. It then updates its speed and angle parameters to change its course of movement from time $t=\frac{T}{2}$ onward so that it can catch Red at time $t=T$. Let $(V^B_2, \Theta_2^B)$ denote the speed and angle parameters that Blue chooses in stage 2. In choosing $V^B_2$, Blue must satisfy the constraint $V^B_1 + V^B_2 \le \overline{V}$, where $\overline{V}$ is a predetermined upper limit. This condition is imposed to reflect constraints such as limited fuel, engine power, etc that an aircraft faces over the course of an actual pursuit maneuver \cite{isaacs1999differential}. Limitations of this sort call for far-sighted upfront decisions in Blue's motion planning so that it stands a reasonable chance of catching Red before it destroys Blue's territory. 

\subsection{Problem Formulation}\label{sec:formul}
In this paper, we are interested in addressing the path planning problem that Blue faces in pursing Red. We aim to design action policies that Blue can consult to optimally choose its speed and angle at each stage of the game. Let $\sigma_i$ denote the stochastic action policy that generates the speed and angle pair $(V^B_i, \Theta^B_i)$. From now on, we refer to $(V^B_i, \Theta^B_i) =: B_i$ and $(V^R_i, \Theta^R_i) =: R_i$ as the \textit{action} taken by Blue and Red, respectively, in the $i$th stage of the game ($i=1,2$).

\subsubsection{Information Structure and Action Policies}
Let $h_i$ denote all the information that Blue knows at the beginning of stage $i$. We call $h_i$ the history at stage $i$ which is given as:
\begin{align}
h_1 &:= \{ r_1 \} \; \; , \;\; h_2 := \{ r_{1:2}, b_1 \} \; , \label{eq:h1} 
\end{align}
where $r_i = (v^R_i,\theta^R_{i})$ and $b_i = (v^B_i, \theta^B_i)$ denote the \textit{actions chosen} by Red and Blue, respectively, in the $i$th stage of the game.
Let $\mathcal{H}_i$ denote the set of all possible values of $h_i$. For any $h_1$, $\sigma_1(h_1) \in \Delta(\mathcal{B}_1)$ is represented by the conditional probability density function (pdf) $\mathbb{P}(B_1 | h_1 ) =: \pi_1(\cdot | h_1)$ under which action $b_1$ is chosen in stage $i=1$ with probability $\pi_1( b_1 | h_1)$. Similarly, given the history $h_2$, $\sigma_2(h_2 ; \overline{V}) \in \Delta(\mathcal{B}_2)$ is represented by the conditional pdf $\mathbb{P}(B_2 | h_2, \overline{V}) =: \pi_2(\cdot | h_2, \overline{V})$ under which action $b_2$ is chosen in stage $i=2$ with probability $\pi_2( b_2 | h_2, \overline{V})$. $\mathcal{B}_i := \mathcal{V}^B_i \times \Phi^B_i$ denotes the set of admissible actions for Blue in stage $i$, where $\mathcal{V}^B_i$ and $\Phi^B_i$ are the sets of admissible speed and angle values, respectively, for Blue in stage $i$. 

\begin{remark}\label{rem:stochPolicy}
\normalfont 
Modeling Blue's action policies $\sigma_{1:2}$ as \textit{stochastic} policies is without loss of generality as they subsume deterministic ones. If the optimal action policy for Blue turns out to be deterministic, it will manifest in the probability with which the optimal action is played under the emergent stochastic policy: the optimal action will be preferred infinitely higher than the sub-optimal ones. 
Furthermore, given that Blue is faced with continuous action spaces $\mathcal{V}_i^B\times \Phi_i^B, i = 1, 2$ containing infinitely many actions, modeling $\sigma_{1:2}$ in the form of probability distributions is more suitable \cite[Chapter 13]{sutton2018reinforcement}. 
\end{remark}

Let $\bar{h}_2 := \{b_1, r_1\}$ denote the set of all the information that Red knows at the beginning of stage $i=2$. 
Let $\overline{\mathcal{H}}_2$ denote the set of all possible values of $\bar{h}_2$. Let the mapping $\rho : \overline{\mathcal{H}}_2 \longrightarrow \mathcal{R}_2$ denote the action policy that Red consults to choose action $R_2$. That is, for any $\bar{h}_2$, $\rho(\bar{h}_2)$ outputs some action $r_2 \in \mathcal{R}_2$, where $\mathcal{R}_2$ denotes the set of admissible actions for Red in stage $i=2$. 


\subsubsection{Blue's Decision Problem}
 Blue's objective is to get as close to Red as possible at time step $t=T$,\footnote{Ideally, Blue wants to catch Red at time $t=T$, i.e., plan its movement such that exactly $x^B_{T} = x^R_{T}$ and $y^B_{T} = y^R_{T}$.} while satisfying its speed upper bound constraint. Blue's problem can then be formulated as follows
\begin{align}\label{eq:Bprob}
\min\limits_{\sigma_{1:2}} \;\;\; &\mathbb{E}\Big\{ \; \sqrt{(X^B_{T} - X^R_{T})^2 + (Y^B_{T} - Y^R_{T})^2} \;\Big\} \notag \\
\text{subject to} \;\; \; &V^B_1 + V^B_2 \le \overline{V},
\end{align}
where for $a \in \{R,B\}$
\begin{align}
X^a_{T} &= X^a_0 + \frac{T}{2} \: V^a_1 \: \cos(\Theta^a_1) + \frac{T}{2} \: V^a_2 \: \cos(\Theta^a_2), \label{eq:X} \\
Y^a_{T} &= Y^a_0 + \frac{T}{2} \: V^a_1 \: \sin(\Theta^a_1) + \frac{T}{2} \: V^a_2 \: \sin(\Theta^a_2), \label{eq:Y}
\end{align}
in which realization of the pair $(V^B_i,\Theta^B_i) = B_i$ is output by the stochastic action policy $\sigma_i(\cdot)$ as described above. For Red, realization of the pair $(V^R_2,\Theta^R_2) = R_2$ is output by the action policy $\rho(\cdot)$. According to the description of the two-step game, $\rho$ can be characterized in terms of the solution to the following functional optimization:
\begin{align}\label{eq:rho*}
\rho := \argmax\limits_{f} \;\; &\sqrt{ (X^R_{T} - X^B_{\frac{T}{2}})^2 + (Y^R_{T} - Y^B_{\frac{T}{2}})^2} \\
\text{subject to} \;\; & \;\; (X^R_{T},Y^R_{T}) \; \text{on Red's safety circle}, \notag
\end{align}
where $(X^R_{T}, Y^R_{T})$ are given by (\ref{eq:X})-(\ref{eq:Y}) in which $(V^R_2, \Theta^R_2) = R_2$ are output by the mapping $f$ that is being optimized in (\ref{eq:rho*}).

\begin{remark}\label{rem:statrho}
\normalfont 
Note that Red can be viewed as a \textit{stationary} agent \cite{hernandez2017survey}, \cite{bucsoniu2010multi} in the sense that its \textit{strategy} for planning its course of movement remains unchanged in response to Blue's actions: its action policy $\rho$ for deciding $(V^R_2, \Theta^R_2)$ is always determined from the optimization problem in (\ref{eq:rho*}). 
\end{remark}

The pair of parameters $(V^R_1,\Theta^R_1) = R_1$ that Red uses to initiate the game are drawn at random from some probability distribution with predetermined statistics in a way that would direct Red towards Blue's territory. 
\begin{assum}\label{assum:rho}
The policy $\rho$ is not known to Blue.
\end{assum}

From Equations (\ref{eq:Bprob})-(\ref{eq:Y}) it is clear that Blue's objective in (\ref{eq:Bprob}) depends on its choice of the parameters $(V^B_1,\Theta^B_1),(V^B_2,\Theta^B_2)$ encapsulated in actions $B_1$ and $B_2$. Let $J(B_{1:2})$ denote Blue's objective in (\ref{eq:Bprob}).

\section{Solution Method}
In this section we develop a solution method for addressing Blue's decision making problem in (\ref{eq:Bprob}).
\subsection{Dynamic Program}
Given that Blue faces a sequential decision making problem under uncertainty, the optimal action policies $\sigma_{1:2}^*$ can be characterized as the solution to a dynamic program with the value functions given below:
\begin{align}
U_2(h_2) &:= \min\limits_{\sigma_2(h_2;\overline{V}) \in \Delta(\mathcal{B}_2)} \;\; \mathbb{E}\Big[ \; J(b_1, \sigma_2(h_2;\overline{V})) \; | \; h_2 \;\Big] \notag \\
&=: \min\limits_{\sigma_2(h_2;\overline{V}) \in \Delta(\mathcal{B}_2)} \;\; \nu_2(h_2, \sigma_2(h_2;\overline{V})) \label{eq:R2} \\
U_1(h_1) &:= \min\limits_{\sigma_1(h_1) \in \Delta(\mathcal{B}_1)} \;\; \mathbb{E}\Big[ \; U_2(H_2) \; | \; h_1 \;\Big] \notag \\
&=: \min\limits_{\sigma_1(h_1) \in \Delta(\mathcal{B}_1)} \;\; \nu_1(h_1, \sigma_1(h_1)), \label{eq:R1}
\end{align}
where $H_2$ is the collection of random variables $\{ R_{1:2}, B_1 \}$. The functions $\nu_1(\cdot)$ and $\nu_2(\cdot)$ are referred to as \textit{cost-to-go} in stages 1 and 2, respectively.

Assumption \ref{assum:rho} implies that Blue does not know the function form of the mapping $J(\cdot)$. Therefore, the optimal policies $\sigma_{1:2}^*$ cannot be found analytically through solving the dynamic program in (\ref{eq:R2})-(\ref{eq:R1}) simply because they are not well-defined from Blue's perspective. 

While Blue's problem cannot be solved analytically, if we have data samples containing motion-associated parameters (i.e., \textit{actions}) chosen by Red and Blue in numerous game scenarios, Blue's policy optimization problem can then be resolved through a \textit{data-driven} approach. Given a sufficiently rich and diverse data set of numerous game scenarios played by Red and Blue, Red's action policy $\rho$ can be learned from those sample game scenarios. Given the stationarity of Red as an agent (see Remark \ref{rem:statrho}), once a suitable learning model has captured and encoded a sufficiently accurate representation of $\rho$ through exposure to many data sampels, it can be used to find the optimal action policies $\sigma_{1:2}^*$ for Blue. 

In this paper, we explore a generative machine learning approach to address Blue's decision making problem through finding implicit representations of $\sigma_{1:2}^*$. Several paradigms are developed for constructing and optimizing generative models \cite{goodfellow2016nips}. Variational Autoencoders (VAEs) \cite{kingma2013auto}, \cite{rezende2014stochastic} and Generative Adversarial Networks (GANs) \cite{goodfellow2014generative} are among the most popular frameworks. In this work, we design and train GAN architectures as decision models that \textit{implicitly} learn representations of the probability distributions underlying the optimal action policies $\sigma_{1:2}^*$. These optimized GAN-based decision models can then be queried to output the optimal choice of actions $B_1$ and $B_2$. We use a large (synthetic) data set containing numerous scenarios of the described two-stage pursuit-evasion game played by Red and Blue. We use the feature values associated with each game instance to train, optimize and evaluate the constructed GAN architectures. 

Since the optimal policies $\sigma^*_{i}, i = 1, 2$ take the form of \textit{conditional} probability distributions, denoted as $\pi^*_1(\cdot | h_1), \pi^*_2(\cdot | h_2, \overline{V})$, we borrow the \textit{class-conditional} implementation of GANs proposed in \cite{mirza2014conditional} and \cite{gauthier2014conditional}. We input Blue's observations to the generative model in order to properly direct the actions output by the model. Additionally we consider inputting other information features into the generative model, such as the upper bound on the speed values for Blue (i.e., $\overline{V}$) in order to properly preserve the \textit{coupling} between Blue's decisions in the two stages of the game. In sequel, we describe our data set and the structure of the generative models that we constructed.



\subsection{Simulated Trajectories}\label{sec:data}
We use a synthetic data set that contains 15000 instances of the described two-stage pursuit-evasion game played between Red and Blue. Each instance of the game is recorded with its associated feature values in $\{(v^R_1, \theta^R_1),(v^B_1, \theta^B_1),$ $(v^R_2, \theta^R_2),(v^B_2, \overline{V} - v^B_1, \theta^B_2), d_{RB}\}$, where $d_{RB}$ denotes the Euclidean distance between the end-points of Red's (i.e., $(X_{20}^R , Y_{20}^R)$) and Blue's (i.e., $(X_{20}^B , Y_{20}^B)$) corresponding trajectories traveled in that instance (see the objective in (\ref{eq:Bprob})). $\overline{V} - v^B_1$ is the speed upper bound for $v^B_2$ in the given game instance. 

To generate each pair of trajectories for the two players, the values of the parameters $(V_1^R, \theta^R_1)$ that initiate a game instance, are drawn at random from a pair of Gaussian distributions $\mathcal{N}(\bar{v}^R, \sigma^R_v)$ and $\mathcal{N}(\bar{\theta}^R, \sigma^R_{\theta})$, where $\bar{v}^R$ and $\bar{\theta}^R$ are the mean values and, $\sigma^R_v$ and $\sigma_{\theta}^R$ denote the standard deviations. The values of the remaining parameters $(v^B_1, \theta^B_1),(v^R_2, \theta^R_2),(v^B_2, \overline{V} - v^B_1, \theta^B_2)$ corresponding to each random draw of the pair $(V_1^R, \theta^R_1)$ were simulated according to the description of the two-stage pursuit-evasion game at the beginning of Section \ref{sec:setup}. 

For the particular data set generated to carry out the experiments presented in the current paper, the statistics of the Gaussian distributions as well as the upper bound $\overline{V}$ were selected such that around 88$\%$ of the generated game instances constituted a successful pursuit maneuver for Blue, i.e., resulting in end-point distances $d_{RB} = 0$. The simulation setup used to synthesize the game instances for the experiments discussed in this paper is outlined in Table \ref{table:simulSetup}. 
\begin{table}
\begin{center} \scalebox{.85}{
 \begin{tabular}{|c|c|c|c|c|c|c|}
 \hline
 $\bar{v}^R$ [spatial units / time step] & $\sigma^R_v$ & $\bar{\theta}^R$ [degrees] & $\sigma^R_{\theta}$ & $\overline{V}$ & $\delta$ & $T$ \\ \hline
  5 & 0.7 & 60$^{\circ}$ & 8$^{\circ}$ & 12 & 10 & 20 \\
 \hline
 \end{tabular}}
 \caption{Simulation setup.}
 \label{table:simulSetup}
 \end{center}
 \end{table}

Figure \ref{fig:trajs3step} demonstrates an instance of the described two-stage pursuit-evasion game in three snapshots. In the game instance displayed in Figure \ref{fig:trajs3step}.(a)-(c), Blue successfully catches Red at time $t=20$ without violating the speed constraint $V^B_1 + V^B_2 \le \overline{V}$ (as evident in Figure \ref{fig:trajs3step}.(c)). Figure \ref{fig:trajs3step} contrasts Blue's two-step path planning with a \textit{single-step} path planning where Blue plans its entire path at time $t=0$, without updating its speed and angle parameters in response to the change in Red's course of movement that occurs in stage 2 (see Figure \ref{fig:trajs3step}.(d)). As a result, Blue fails to catch Red at time $t=20$. Instead, Blue ends up at the destination it \textit{initially} projected for Red, i.e., the end-point of the yellow-crossed line. The yellow-crossed line marks Red's trajectory, had it not updated its path at time $t=11$.

\subsection{Generative Model Architectures}\label{sec:model} 
In this work, following the conditional variant of GANs, we construct two generator networks, one for Blue's decision at each stage of the game. Let $G_i$ denote the generator network that is trained to represent the conditional pdf $\pi^*_i$ which characterizes the optimal stochastic policy $\sigma_i^*$. Once trained and optimized, given Blue's observation of $R_i$, generator $G_i$ can be queried to output $b_i$ as the action recommended to Blue in stage $i$. 

The input-output structure of the generative networks that we constructed are shown in Figure \ref{fig:testg12}. Generator $G_1$ (the green block) is fed Blue's observation of $R_1$ as input along with a random draw of the independent Gaussian noise $Z_1$ and a quality score $s_1$. These features altogether are then decoded by the generative network into the action $b_1$ that determines Blue's path in stage 1. The quality score $s_1$ is a measure of the \textit{effectiveness} of action $b_1$ that is output by $G_1$ in response to the observed value of $R_1$. 
In our work, for the training phase we use a normalized version of the Euclidean distance between the end-points of Red's and Blue's corresponding spatial trajectories in each instance of the game, as the quality score $s_1$. Later, we elaborate on the procedure that we use to construct the score $s_1$ for each sample pair of Red and Blue trajectories. 

Generator $G_2$ (the blue block in Figure \ref{fig:testg12}) is fed more information as inputs compared to $G_1$: In addition to the independent latent noise draw $Z_2$ and the observed realizations of $R_1$ and $R_2$, $G_2$ is also fed the upper-bound $(\overline{V}-v^B_1)$ on the stage 2 speed $V^B_2$ so as to capture and preserve the \textit{coupling} between Blue's decisions in the two consecutive stages of the game. 
Moreover, the quality score $s_2$ is fed to $G_2$ so as to direct the generator towards outputting a \textit{highly effective} action $B_2$ for Blue in response to the realized $R_1$ and $R_2$. Similar to $s_1$, the quality score $s_2$ is generated based on a normalized version of the Euclidean distance between the end-points of Red's and Blue's corresponding paths in each game instance. 
More specifically, let $d^i_{RB}$ denote the end-point distance between Red and Blue's corresponding paths in the $i$th game instance in the data set and let $\eta(d^i_{RB}) \in [0,1]$ denote the distance $d^i_{RB}$ \textit{normalized} across the training data set. 0 and 1 are the normalized lower and upper bounds, respectively, on the normalized end-point distance values in the data set. We then define $s_2^i := 1 - \eta(d^i_{RB})$ as the quality score assigned to a pair of Red's and Blue's paths with the end-point distance $d^i_{RB}$.\footnote{Note that quality score $s_2$ computed as such for a pair of Red and Blue trajectories associated with some $h_2, b_2$, is nothing but a normalized version of the \textit{cost-to-go} function $\nu_2(\cdot)$ in (\ref{eq:R2}) evaluated at $h_2, b_2$. Indeed the end-point distance $d_{RB}$ resulting from $h_2, b_2$ equals $\nu_2(h_2, b_2)$.\label{fn:r1s1}} Therefore, the smaller the distance $d^i_{RB}$, the larger the corresponding score $s_2^i$. 
After the model is trained, we can query  generator $G_2$ with high  score values $s_2$ within the normalized $[0,1]$ regime to guide it towards outputting highly effective actions $B_2$ for Blue in stage 2. 

From the dynamic programming formulation in (\ref{eq:R2})-(\ref{eq:R1}) it is clear that the characterization of $\sigma_1^*$ depends on the terminal value function $U_2(\cdot)$ and thus, requires $\sigma_2^*$ to be resolved. Therefore, given that $G_1$ implicitly represents $\sigma_1^*(\cdot)$, its training requires $G_2$ to be trained. 
Given the quality scores $s_2$ computed across the training data samples as described above, we first train generator $G_2$ and then use that to construct the quality scores $s_1$ for all the game instances in the training data set. These scores are then used to train generator $G_1$. Both generators are modeled with fully-connected multi-layer neural networks. 
The setup that we used to construct the GAN architectures in terms of the constituent hyper-parameters is outlined in Table \ref{table:GDnets}.\footnote{The setup outlined in Table \ref{table:GDnets} was chosen after experimentation with several architectures with different numbers of hidden layers, neurons in each layer, etc.} Next, we describe the procedure we designed to construct the quality scores $s_1$.
\begin{table}
\begin{center} \scalebox{.65}{
 \begin{tabular}{|l||c|c|c|}
 \hline 
 \backslashbox{Module}{Hyper-parameters} & $\#$ neurons in HL1 & $\#$ neurons HL2 & Input Noise Dimension \\ \hline\hline
  Generative Network & 96 & 64 & 2 \\ \hline
  Discriminative Network & 64 & 32 & 2 \\
 \hline
 \end{tabular}}
 \caption{GAN Architecture Setting: the generative and discriminative networks each constitute two hidden layers denoted as HL1 and HL2.}
 \label{table:GDnets}
 \end{center}
 \end{table}
 

\subsubsection{Construction of the quality score $s_1$}\label{sec:s1build}
Assuming generator $G_2$ is trained, the procedure that we use to construct the quality scores $s_1$ for the game instances in the data set is outlined in Algorithm \ref{alg:s1scores}. For the $i$th game instance, the values of the features $\{(R_{1:2}),\overline{V}-V^B_1\}_i$ along with the maximum normalized quality score $\bar{s}_2 (\approx 1)$ and the random noise input $Z_2$ are fed to the optimized $G_2$ to generate action $B_2$.
Using a Monte Carlo sampling approach, for each game instance, $G_2$ is queried $N_{MC}$ times (e.g., 30 times) to generate $N_{MC}$ recommendations for $B_2$. The resultant end-point distances $d^j_{RB}$ corresponding to each generated action $(B_2)_j$ are then averaged as $\alpha^i := \frac{1}{N_{MC}}\sum\limits_{j=1}^{N_{MC}} d^j_{RB}$. These averaged quantities $\alpha^i$'s associated with each game instance in the data set, are then stacked and normalized across the entire data set. So that given each $\alpha^i$, $\eta(\alpha^i) \in [0,1]$ gives its normalized value. Then, for the $i$th game instance: $s^i_1 = 1 - \eta(\alpha^i)$. 

The quantities $\alpha^i$ are connected to the cost-to-go function $\nu_1(\cdot)$ in (\ref{eq:R1}) in the same way that the end-point distances $d^i_{RB}$ are connected to $\nu_2(\cdot)$ in (\ref{eq:R2}) (see footnote \ref{fn:r1s1}). Basically, given the feature values $h_1,b_1$ (see (\ref{eq:h1})) associated with the $i$th game instance, the corresponding quantity $\alpha^i$ represents an approximation of $\nu_1(h_1,b_1)$ (see (\ref{eq:R2})). Therefore given the input features $h_1,b_1$, the quality score mapping $s^i_1 = 1 - \eta(\cdot)$ approximately outputs a normalized version of $\nu_1(h_1,b_1)$.


\begin{algorithm} [ht]
 \caption{Pseudocode for construction of $s_1$ scores}
 \label{alg:s1scores}
 \begin{algorithmic}[1]
	\For{$i=1 \; \; \text{to} \; \; |\mathcal{D}|$}
	\State $d^i \; \longleftarrow \; [ \;\; ]$
		\For{$j=1 \;\; \text{to} \; \; N_{MC}$}
			\State query $G_2$ with the input feature values $\{r_{1:2}, \overline{V}-v_1^B\}_i$ associated with the $i$th game instance, $\bar{s}_2 (\approx 1)$ and the latent noise draw $z_2$
			\State Use $(b_2)_j$ \textit{output} by $G_2$, to obtain the corresponding end-point distance $d^j_{RB}$ between the resultant Red and Blue trajectories
			\State $d^i \; \longleftarrow \; [\; d^i \; , \; d^j_{RB} \;]$ 
		\EndFor
		\State $\alpha^i \; \longleftarrow \; $ average of the entries in $d^i$ vector 
		\State $s^i_1 \; \longleftarrow \; 1 - \eta(\alpha^i)$
	\EndFor
\State \textbf{return} $\; \{s^i_1\}_{i=1}^{|\mathcal{D}|}$
 \end{algorithmic}
\end{algorithm} 

\section{Evaluation of the Sequential Decision Architecture}\label{sec:test}
\begin{figure}[h]
\centering
\includegraphics[scale=.25]{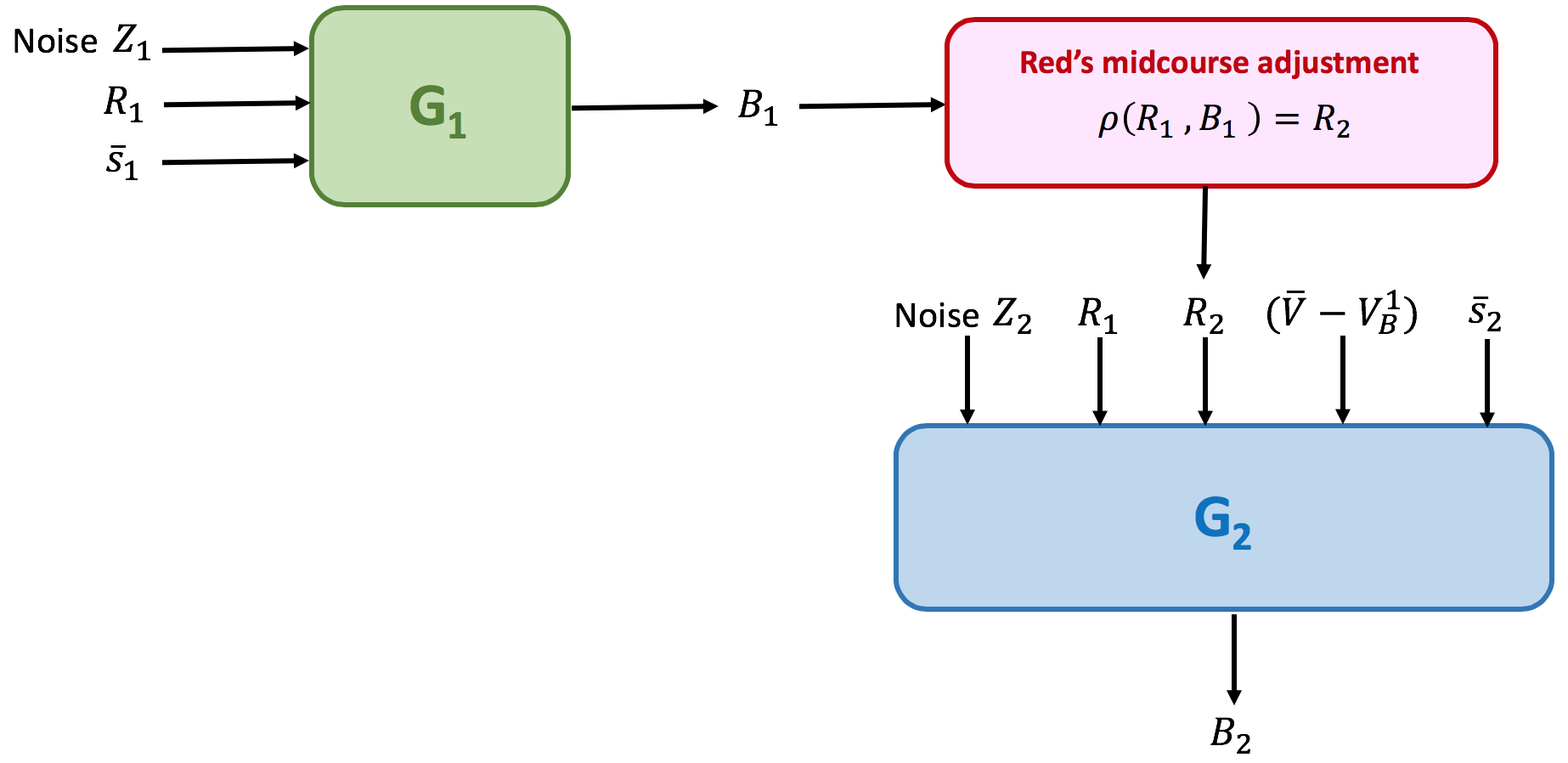}
\caption{Illustration of the procedure used to test the two-step decision model.}
\label{fig:testg12}
\end{figure}

After $G_1$ and $G_2$ are trained, they can be queried sequentially to generate the actions $B_1$ and $B_2$, respectively. The evaluation procedure is illustrated in Figure \ref{fig:testg12}. For each $R_1$ that initiates a game instance, $G_1$ is queried with the realized $R_1$ and the score $\bar{s}_1 (\approx 1)$, as well as a random noise draw $z_1$. The action $B_1$ output by $G_1$ is then used to plan Blue's spatial trajectory in stage 1 in response to Red's path induced by the realized $R_1$. The action $R_2$ that is chosen by Red in stage 2 is recovered (see the pink block in Figure \ref{fig:testg12}) from Red's action policy $\rho$ (see (\ref{eq:rho*})). Action $R_2$ that is observed by Blue in stage 2 is then fed to $G_2$ along with the realizations of the other input features (see Figure \ref{fig:testg12}). 
 $G_2$ outputs the action $B_2$ which is used by Blue to plan its path in stage 2 in response to Red's updated trajectory. The resultant end-point distance $d_{RB}$ can then be calculated according to the Euclidean distance metric in (\ref{eq:Bprob}). These distances are then aggregated across all the game instances in the test data set to statistically assess the performance of the decision model. 

\subsection{Decision Benchmarks}\label{sec:results}
We considered two different benchmark decision models to assess the performance of our proposed two-step decision model against them. We describe these benchmarks below.

\textbf{Single-step decision benchmark:} Under this benchmark decision policy, Blue plans its \textit{entire trajectory} based solely on its initial observation of $R_1$. This benchmark, too, is constructed using the GAN model, but it consists of a \textit{single generator} that is queried with the observed value of $R_1$ in stage 1 as well as the random noise input.\footnote{The generator in this GAN-based decision benchmark was trained using a procedure very similar to the one we used to train $G_1$ and $G_2$ (see Section \ref{sec:model}). } It then recommends action $B$ that is used by Blue to plan its entire trajectory in the form of a single line over the course of the two stages of the game. Under this single-step benchmark, observation of $R_2$ is not incorporated to adjust Blue's trajectory in stage 2. Comparison of our two-step decision model against this benchmark would thus shed light on the effect of incorporating Blue's mid-course observation of $R_2$ in the quality of its \textit{long-range} path planning. 

An example of the pursuit-evasion game where Blue continues on a single path that it plans based solely on its initially decided speed and angle parameters $(V,\Theta)$ is shown in Figure \ref{fig:trajs3step}.(d). 
Note that Blue still faces the speed constraint $V^B_1 + V^B_2 = 2 V \le \overline{V}$. Hence if $2V > \overline{V}$, Blue's speed in stage 2 drops to $\overline{V} - V$.

\textbf{Randomized two-step decision benchmark:} To demonstrate the effect of using GAN architectures in constructing the decision model, we also consider a pair of randomized decision policies with statistics inferred from the training data set. More specifically, under the randomized decision benchmark, the pair of speed and angle parameters $(V^B_i, \Theta^B_i)$ that constitutes action $B_i$, are drawn from a two-dimensional Gaussian distribution $\mathcal{N}(\ubar{\mu}^i,\Sigma^i)$ where, $\ubar{\mu}^i := [\mu^i_v,\mu^i_{\theta}]^T$ denotes the mean vector and $\Sigma^i$ denotes the covariance matrix. The statistics of these Gaussian distributions are given by the corresponding sample statistics computed based on the samples in the training data set. 

\subsection{Generated Trajectories: Examples}\label{sec:assess}
\begin{figure*}[!ht]
\centering
\includegraphics[width=2\columnwidth, keepaspectratio]{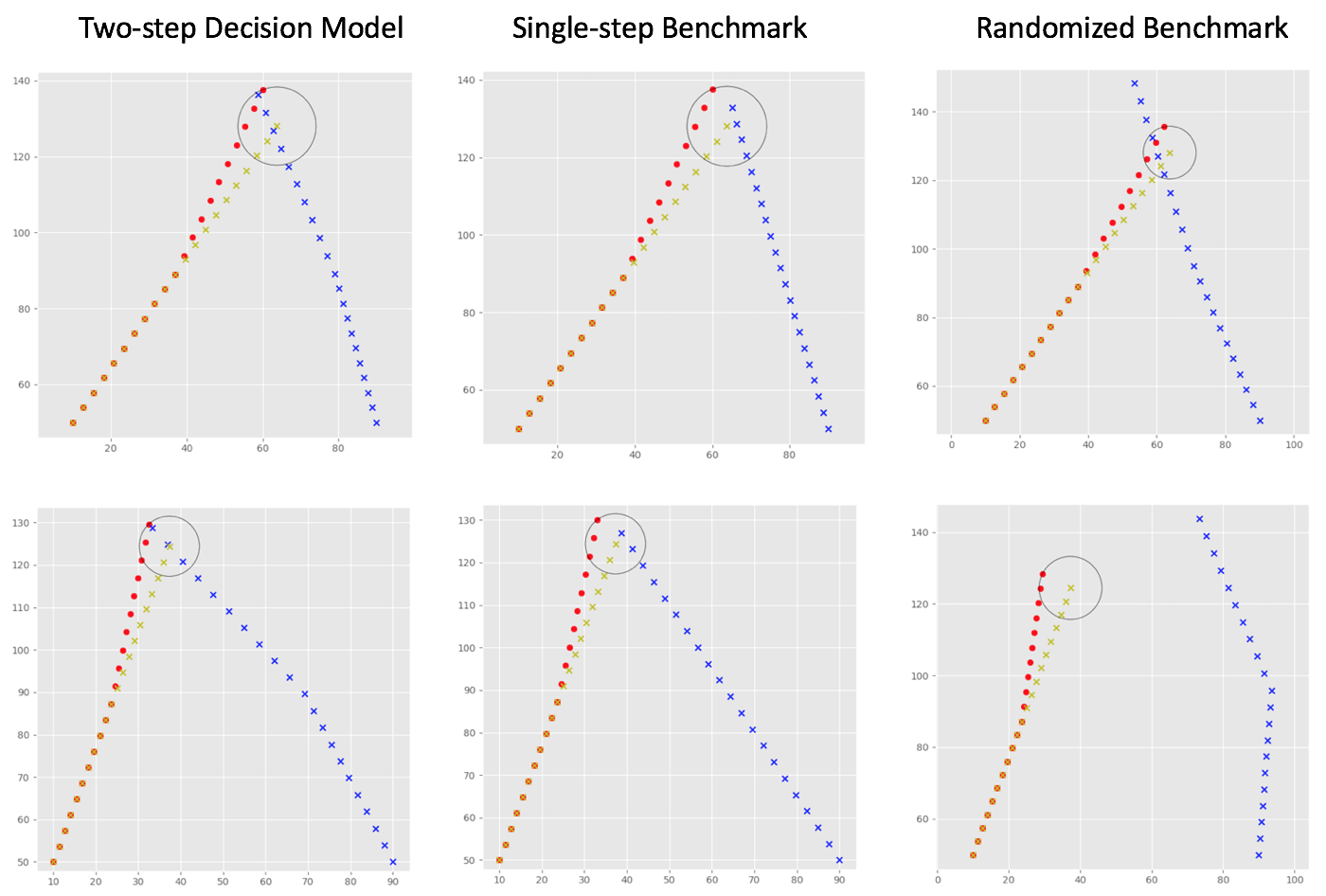}
\caption{Examples of Red's and Blue's corresponding trajectories that emerged under the actions generated by the two-step decision model (left column), single-step (middle column) and randomized (right column) decision benchmarks. In the images in each row, Red initiates the game with the exact same realization of $R_1$. } 
\label{fig:trajs}
\end{figure*}

Figure \ref{fig:trajs} demonstrate two sets of pursuit-evasion game scenarios (one per row) that emerge under our proposed two-step decision model (left column), single-step (middle column) and the randomized (right column) decision benchmarks. In each of the three images shown within each row in Figure \ref{fig:trajs}, the game is initiated by the same realized action $R_1$. Therefore, we can visually compare how Blue responds to the same Red's path in stage 1 under each of the three decision models.
 In these examples it is visually evident that the path generated for Blue using the actions recommended by the two-step decision model has enabled Blue to get much closer to Red's destination at time $t = 20$ compared to the benchmarks. We observe that the Blue trajectories generated under the two-step decision model \textit{almost coincide} with Red's paths at time $t = 20$. However, the trajectories that result under the benchmarks end up at points much farther from the end-point of Red trajectory. 

\subsection{Statistical Assessment}\label{sec:statTest} 
To statistically evaluate the performance of our proposed decision architecture, we generated spatial trajectories for Blue using the actions recommended by the two-step decision model, in response to Red's initiated paths in all the game instances in the test data set which contained 3750 samples. For each game instance we also generated trajectories for Blue using the single-step and randomized decision benchmarks. For each resultant pair of Red and Blue trajectories under each of these three decision models, we recorded the corresponding end-point distances between the paths. 

We computed the \textit{difference} between the end-point distance values that result under the two-step model and the two benchmarks, \textit{on each} game instance. Let $d^*_i$ denote the end-point distance between Red and Blue trajectories that emerge in the $i$th game instance under the two-step model. Let $d^s_{i}$ and $d^r_{i}$ denote the end-point distances between Red and Blue trajectories in the same game instance, that emerge under the single-step and randomized decision benchmarks, respectively. Define $\delta^{*s}_i := d^*_i - d^s_i$ as the difference between the end-point distances that result under the two-step model ($d^*_i$) and the single-step benchmark ($d^s_i$) in the $i$th game instance. 
In a similar fashion define and interpret the distance measures $\delta^{*r}_i := d^*_i - d^r_i$ and $\delta^{sr}_i := d^s_i - d^r_i$ for the $i$th game instance.

\begin{figure}[!h]
\centering
\includegraphics[scale=.55]{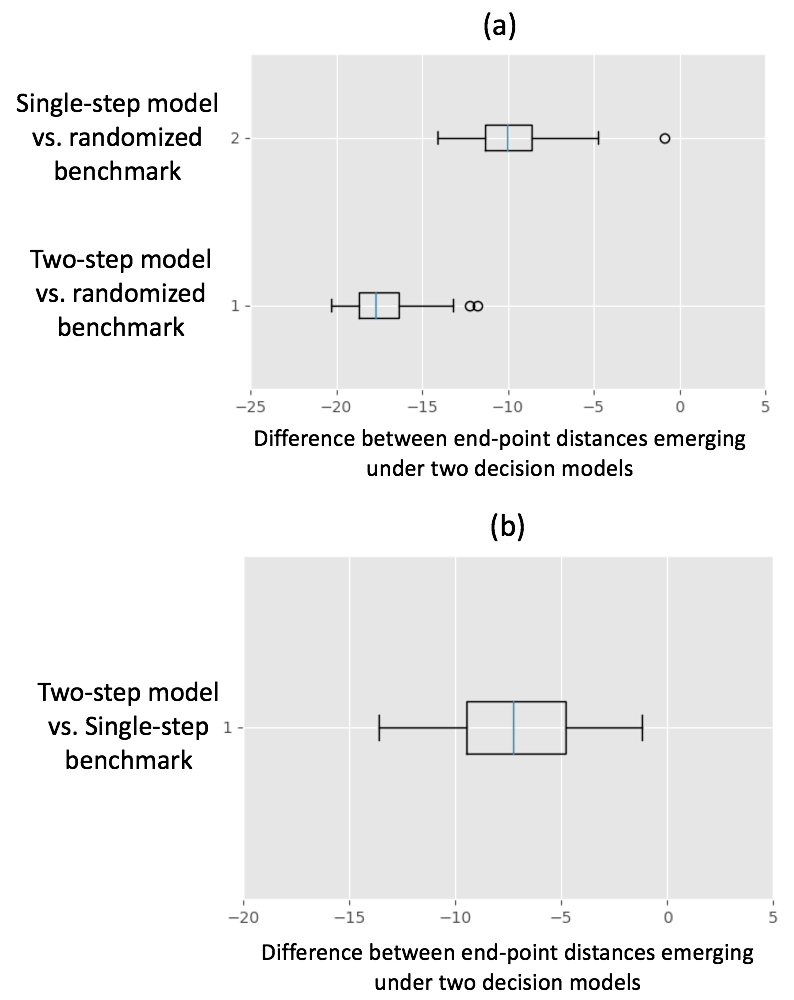}
\caption{Box plots of the \textit{summary} statistics (sample averages) of the difference between each pair of models, in terms of the resultant end-point distances. (a) corresponds to the average difference between the GAN-based models and the randomized benchmark. (b) corresponds to the average difference between the two-step decision model and the single-step benchmark. Notice that \textbf{all the resultant values are \underline{negative}}. This means that both GAN-based models outperform the randomized benchmark (a) and, the two-step decision model is superior to the single-step benchmark (b). The plots are based on the evaluation of 50 separately trained instances of the two-step decision model. }
\label{fig:summaryStat_boxes}
\end{figure}

We trained 50 instances of the two-step decision model and evaluated each of them separately. In each of these experiments $75 \%$ of the samples were used for training while $25 \%$ were used for testing. While testing each trained instance of the two-step decision model, we computed the values of the three measures $\delta^{*s}_i, \delta^{*r}_i$ and $\delta^{sr}_i$ across all the 3750 game instances in the test data set, and stored their \textit{sample-mean values} for each of the trained 50 instances of the two-step decision model.
 These sample-mean values are box plotted in Figure \ref{fig:summaryStat_boxes}. 

Box plots of the average $\delta^{*r}$ and $\delta^{sr}$ quantities are displayed in rows 1 and 2, respectively, in Figure \ref{fig:summaryStat_boxes}(a). We see that all the resultant values including the few outliers are \underline{negative}. This implies that on an aggregate level both two-step and single-step decision models which are GAN-based architectures, outperform the randomized decision benchmark in terms of the resultant end-point distances between Red and Blue trajectories. Figure \ref{fig:summaryStat_boxes}.(b) displays the box plot corresponding to the average $\delta^{*s}$ quantities. Again all the resultant values are \underline{negative} which indicates the superiority of the two-step decision model to the single-step benchmark, in terms of the end-point distance between Red and Blue trajectories. These assessments support the idea that incorporating Blue's mid-course observations in adaptively planning its trajectory in response to alterations in Red's path leads to more effective paths for Blue in terms of their ultimate closeness to Red's destination. 

\subsection{Speed Constraint Satisfaction}
Recall that Blue faces a constrained optimization problem in planning its trajectory: sum of its speed values in stages 1 and 2 cannot exceed the upper bound $\overline{V}$ (see (\ref{eq:Bprob})). Therefore, it would be interesting to examine whether the speed values output by the GAN-based decision architectures satisfy this constraint. This would then demonstrate the capability of these GAN-based models in implicitly learning and encoding constraints of this sort. To investigate this property, we recorded the number of speed constraint violations that occurred under each decision model when evaluating them on the test data set (3750 samples). 
The results are outlined in Table \ref{table:Vviol}. We observe that the two-step decision model has resulted in the highest number of speed constraint violations among the three decision models. Whereas, the single-step decision benchmark has resulted in the fewest of such violations with relatively significant difference. The number of constraint violations being the highest under the two-step decision model, might be attributed to its greater model complexity. Recall that the two-step decision model consists of two generators that are queried in a sequential fashion (see Figure \ref{fig:testg12}). Therefore, the number of input features as well as the number of constituent neural networks in the structure of the two-step model is much larger compared to those in the single-step model which consists of a single generator. Nevertheless, the number of speed constraint violations is relatively small in all three models, compared to the size of the test data set (e.g., $\approx 6.6 \%$ of the test data samples, under the two-step model). Therefore, in practice the decision model can be queried multiple times to obtain feasible recommendations for the speed parameter. 
 This observation is solely reflective of the data set that we used to conduct these experiments. Further studies are needed in order to achieve more conclusive observations and insights about the relation between the model complexity and implicit learning of the problem constraints. 
\begin{table}
\begin{center} \scalebox{1}{
 \begin{tabular}{|l|c|}
 \hline
 Decision Model & $\#$ Speed constraint violations \\ \hline \hline
 Two-step decision model & 247 \\ \hline
 Single-step decision benchmark & 46 \\ \hline
 Randomized decision benchmark & 102 \\ 
 \hline
 \end{tabular}}
 \caption{Speed constraint violations occurred under each decision model.}
 \label{table:Vviol}
 \end{center}
 \end{table}

\subsection{Quality Scores Sensitivity Analysis}\label{sec:score}
\begin{figure*}[!ht]
\centering
\includegraphics[width=2\columnwidth, keepaspectratio]{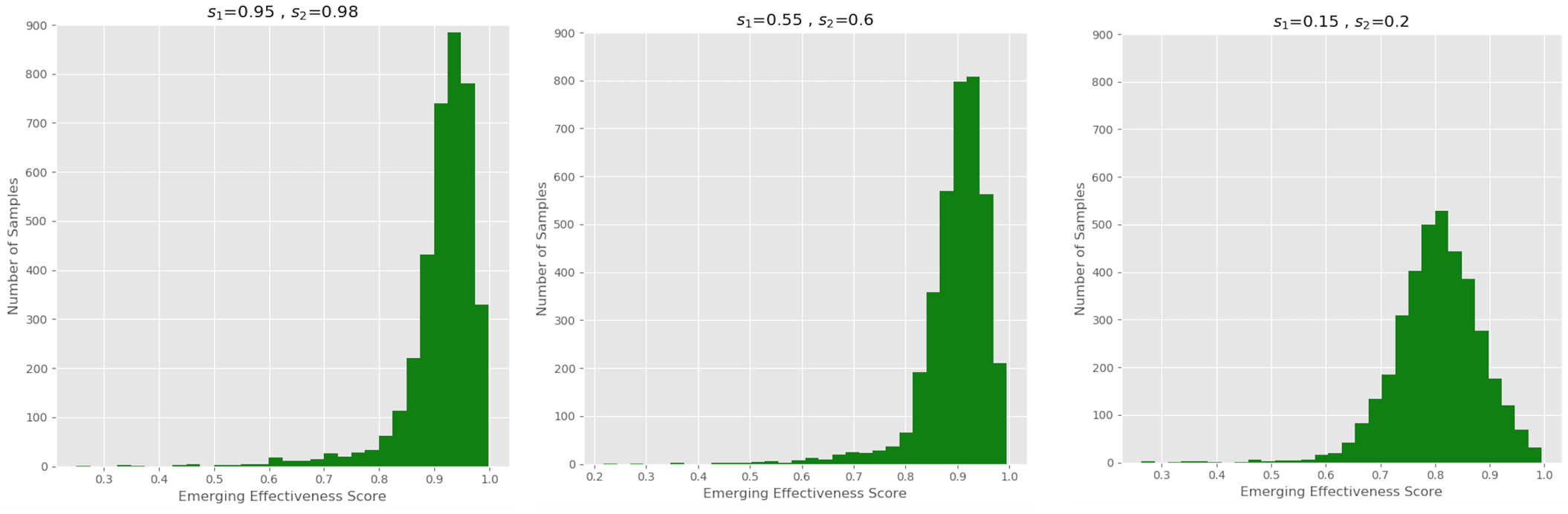}
\caption{Histograms of the \textit{emerging} quality scores resulting under three different settings of the quality score values $s_1$ and $s_2$ fed to the two-step decision model. The histogram skewness becomes further negative as the input values for $s_1$ and $s_2$ increase: the higher the input scores, the smaller the resultant end-point distances between the emerging Red and Blue trajectories. }
\label{fig:Oscores_hist}
\end{figure*}

As discussed in Section \ref{sec:model}, the quality scores $s_1$, $s_2$ fed to the two-step decision model as part of the input features (see Figure \ref{fig:testg12}) are intended to provide some control over the effectiveness of the actions recommended by the model. Recall that in our design, higher score values within the regime $[0,1]$ are indicative of better trajectories for Blue. Therefore, we expect the model to give rise to more effective trajectories for Blue once queried with higher values of $s_1$ and $s_2$. To test this hypothesis, we queried the two-step decision model with various settings for $s_1,s_2$ values. For each setting we contrasted the \textit{input} values for $s_1,s_2$ with the quality score computed for the \textit{emerging} Red and Blue trajectories. More precisely, let $d^i(s_1,s_2)$ denote the end-point distance between the Red and Blue trajectories that emerge when the model is queried with $s_1,s_2$ values. Let $\eta(d^i(s_1,s_2)) \in [0,1]$ denote the \textit{normalized} version of the distance $d^i(s_1,s_2)$, where the normalization carried out via $\eta(\cdot)$ is with reference to the end-point distance values in the training data set. We then use $1-\eta(d^i(s_1,s_2))$ as the \textit{realized} quality score associated with the $i$th game instance in the test data set. Figure \ref{fig:Oscores_hist} shows the histograms of these realized quality scores under three different settings for $s_1$ and $s_2$ input values. We observe that on an aggregate level, these realized scores are relatively correlated with the input values for $s_1$ and $s_2$: the higher the input score values, the higher the realized quality score. As can be seen, under the settings $s_1 = 0.95, s_2 = 0.98$ and $s_1 = 0.55, s_2 = 0.6$ the great majority of the realized quality scores lie above 0.8, with the skewness being further negative under the former. Whereas, the smaller input score values $s_1 = 0.15, s_2 = 0.2$ have resulted in much poorer realized scores: nearly half of the emerging game scenarios in the test set have scored \textit{below} 0.8 in this case.

The observations we made based on Figure \ref{fig:Oscores_hist} imply that the input features $s_1,s_2$ provide some control over the quality of the trajectories that the two-step model gives rise to. By inputting higher score values in the normalized regime $[0,1]$ we can prompt the two-step model to generate better trajectories for Blue, i.e., ones that enable Blue to get closer to Red's destination by time $t=20$.

\section{Conclusion}\label{sec:conclusion}
We studied a two-stage pursuit-evasion game played between two agents, Blue (the pursuer) and Red (the evader), over $T=20$ time steps. 
We focused on the decision making problem that Blue faces, i.e., choosing its motion-associated parameters (speed and angle) such that it can catch Red at time $T$. 
We developed generative model architectures to implicitly learn and represent Blue's optimal action policy for deciding its speed and angle parameter values in each stage of the game. 
In particular, we constructed GAN architectures consisting of two generative networks that need to be queried in a sequential fashion to generate the optimal actions for Blue in each stage of the game. 
We devised a scoring system that can be used to control the quality of the actions output by the model, in terms of the effectiveness of the pursuit path they give rise to for Blue. 
We conducted various statistical assessments to demonstrate performance of the developed framework. 
The obtained results suggest that generative modeling methods offer considerable potential for addressing the strategic long-range decision making problems based on data-driven approaches. 

Extending the present study to consider pursuit-evasion game scenarios in presence of more sophisticated path-planning constraints beyond the speed limitations, is an interesting future direction. 
Moreover, investigating pursuit-evasion games under an infinite time-horizon is an important extension to the present work that needs to be explored in future. 
In addition, experimenting with other generative modeling frameworks such as VAEs would provide further insights into the effectiveness of generative machine learning paradigm for policy optimization in multi-agent planning problems.

Finally, the decision models discussed in this paper implicitly incorporate approximations of the Red agent's behavior. 
Modeling adversary behavior in a game is referred to as \textit{opponent modeling}.
Our work is particularly related to \textit{implicit} opponent modeling \cite[Section 4.8.1]{albrecht2018autonomous} where certain aspects of the opponent's strategy or behavior are implicitly encoded in various representation forms to be used for downstream computations such as policy optimization. 
Further work is needed to explore  the development and incorporation of better opponent models in the game discussed here.
For example, He et al. in \cite{he2016opponent} develop a framework where instead of explicitly predicting the opponent's behavior, a hidden representation of the opponent is learned and then used to compute an adaptive response. 
Bard et al in \cite{bard2013online} construct a portfolio of policies offline and then use online learning algorithms to select the best response strategy from the portfolio during online interactions.



\bibliographystyle{plain}
\bibliography{REF}

\end{document}